\documentclass[lettersize,journal]{IEEEtran}
\usepackage{amsmath,amsfonts}
\usepackage{algorithmic}
\usepackage{algorithm}
\usepackage{array}
\usepackage[caption=false,font=normalsize,labelfont=sf,textfont=sf]{subfig}
\usepackage{textcomp}
\usepackage{stfloats}
\usepackage{url}
\usepackage{verbatim}
\usepackage{graphicx}
\usepackage{cite}
\usepackage{tabularx}
\usepackage{booktabs}
\usepackage{dsfont}
\usepackage[colorlinks=true, allcolors=blue]{hyperref}

\begin{document}

\title{Harnessing Attention Mechanisms: Efficient Sequence Reduction using Attention-based Autoencoders}

\author{Daniel Biermann\IEEEauthorrefmark{1}, Fabrizio Palumbo\IEEEauthorrefmark{1}\IEEEauthorrefmark{2}, Morten Goodwin\IEEEauthorrefmark{1}, Ole-Christoffer Granmo\IEEEauthorrefmark{1} \newline\IEEEmembership{\IEEEauthorrefmark{1}Centre for Artificial Intelligence Research, University of Agder, Norway\\ 
Department of ICT, University of Agder, Grimstad, Norway\\ \and 
\IEEEauthorrefmark{2}Artificial Intelligence Lab (AI Lab), Institutt for informasjonsteknologi,\\ 
Oslo Metropolitan University, Oslo, Norway
}
\thanks{}}


\maketitle

\begin{abstract}
Many machine learning models use the manipulation of dimensions as a driving force to enable models to identify and learn important features in data. In the case of sequential data this manipulation usually happens on the token dimension level. Despite the fact that many tasks require a change in sequence length itself, the step of sequence length reduction usually happens out of necessity and in a single step. As far as we are aware, no model uses the sequence length reduction step as an additional opportunity to tune the models performance. In fact, sequence length manipulation as a whole seems to be an overlooked direction. In this study we introduce a novel attention-based method that allows for the direct manipulation of sequence lengths. To explore the method's capabilities, we employ it in an autoencoder model. The autoencoder reduces the input sequence to a smaller sequence in latent space. It then aims to reproduce the original sequence from this reduced form. In this setting, we explore the methods reduction performance for different input and latent sequence lengths. We are able to show that the autoencoder retains all the significant information when reducing the original sequence to half its original size. When reducing down to as low as a quarter of its original size, the autoencoder is still able to reproduce the original sequence with an accuracy of around 90\%.
\end{abstract}


\section{Introduction}
\IEEEPARstart{O}{ver} the recent years, a lot of progress has been made in the field of natural language processing (NLP). This progress has been largely driven by the Transformer, introduced by Vaswani et al. in 2017 \cite{vaswani2017attention}. The power of Transformer models is based on their ability to avoid recurrence in favour of an easy parallelizable attention mechanism while retaining the ability to capture contextual information in sequential data. Since then, many Transformer-based models have been developed, studied, and used to reach ever-better-performing NLP models. Among these are OpenAI's GPT models \cite{radford2018improving,radford2019,NEURIPS2020_1457c0d6}, XLNet \cite{NEURIPS2019_dc6a7e65} and, BERT \cite{devlin-etal-2019-bert} and its popular derivatives \cite{sanh2019distilbert,liu2019roberta}. OpenAI's most recent GPT iterations, ChatGPT \cite{ChatGPT} and GPT4 \cite{openai2023gpt4} have generated a lot of media attention and demonstrated the power and fast progress of such models in NLP.

Nearly all NLP tasks require the model to change the shape of the input sequence at some level in the workflow. The desired output shape rarely corresponds to a sequence's input shape. A core problem in many classification tasks is the reduction of sequences with many tokens down to a single token, as in many classification tasks or sentence embedding. Despite this, few Transformer models deviate from the standard practice of capturing data features solely on the word token level. The sequence reduction is usually performed in a way that does not allow for a lot of exploration and tuning on the sequence level (see. \ref{sec:related_work}). 

The proclivity to capture contextual information and data features on the word token level lies in the attention mechanism used in Transformer models. The scaled dot-product attention \cite{vaswani2017attention} generates a new contextual representation for each token in a sequence by calculating a weighted sum over the tokens in the entire sequence. The weights, which correspond to an attention map between all tokens, yield from generated query, key and value vector representations of each token. These vectors are created from the tokens in the input sequence. Thus, the scaled dot-product naturally captures contextual information on a word token level.

Interestingly, by design, the scaled dot-product attention does allows for direct manipulation of the number of tokens in the sequence, as it was first introduced in a machine translation task \cite{vaswani2017attention}.
The number of tokens output by scaled dot-product attention is dictated by the number of query vectors. While it is clear how to generate differing query vector numbers in a machine translation setting, in general it is not as straightforward how to initialize query vectors so that the number of query vectors differs from the input sequence. Due to this, it seems that in Transformer models, the sequence reduction step itself is done as a necessity and seldomly seen as an opportunity for additional modelling and tuning.   

In general, the investigation and use of more nuanced techniques to manipulate the sequence length is of interest. Manipulating the sequence length could offer an additional axis and tool to tune and create better and more potent latent spaces that capture the patterns in sequential data. Additionally, manipulating sequence length could allow for avoiding or alleviating the problem of empty calculations due to padding sequences or putting more context into input-restrictive models such as BERT and other Transformer models.

In this work, we investigate the ability to use scaled-dot product attention to capture features by directly manipulating the number of tokens in a sequence instead of the dimensions of its tokens. To this end, we introduce an additional scaling matrix into the scaled dot-product to enable it to manipulate the number of tokens in a sequence more freely. We then employ this reducing scaled dot-product attention in an autoencoder setting. The autoencoder encodes sequences in a latent space with fewer tokens and recreates the entire original sentence from the token-reduced latent space.
While our method introduced strong restrictions regarding the uniformity of input sentences, we argue that these restrictions overlap and synergize well with existing input shape limitations of existing popular attention models such as BERT-based models. In particular tasks with very long input sequences will be less affected due to already existing limitations.
Overall, in this paper:
\begin{itemize}
\item We introduce reducing scaled dot-product attention. By adding a simple scaling matrix to the query vector generation process in the scaled dot-product attention process, we enable it to directly manipulate sequence length.
\item We investigate the reducing scaled dot-product's ability to retain information when reducing sequence lengths for different reduction sizes and input sequence lengths.
\item We build and train a novel attention-based autoencoder that creates a latent space by manipulating sequence dimensions instead of token dimensions.
\end{itemize}
To the best of our knowledge, there is currently no other detailed work investigating the direct manipulation of sequence lengths with Transformer-like attention mechanisms. Further, we are unaware of other investigations using nuanced sequence length manipulation as a primary tool to encode contextual information in latent spaces.

Thus, we hope that the explorations in this work spark inspiration in other researchers to explore the manipulation of sequence lengths in addition to token dimension. We are further convinced that, in time, more investigations in this direction will reveal more natural and less restrictive methods to directly manipulate sequence length with attention.

\section{Related Work}
\label{sec:related_work}
When handling sequential data, most machine learning tasks require the model to change the shape of the input to the shape of the desired output. The desired output can range from a static number of classes in classification tasks to a new sequence of different shape in text generation tasks to a single token in sentence embedding tasks. This sequence reduction step is performed in different ways depending on the model.

In recurrent models, the sequence length reduction is achieved in tandem with the context capturing mechanism. Gated-recurrent Unit (GRU) \cite{gru} models or long short-term memory (LSTM) \cite{lstm} models process the information sequentially and the last hidden state captures the contextual information of the entire sequence. While naturally reducing a sequence down in length, difficulties in parallelization and optimization make these models challenging in their own regard \cite{6639349}.

The majority of current state-of-the-art models in NLP are based upon the attention mechanism introduced with the Transformer model \cite{vaswani2017attention}. BERT and the GPT models make use of the basic Transformer encoder/decoder blocks, while other models keep the overall structure of a Transformer model and replace the scaled dot-product and multi-head attention mechanisms with more efficient attention models. For example, the Longformer \cite{beltagy2020longformer} replaces the scaled-dot product with an attention mechanism that scales better with the length of an input sequence.
Unlike in recurrent models, the attention mechanism does not naturally yield a reduction down to one token for classification tasks. While the mechanism itself theoretically allows to change the length of the sequence during the attention step, in practice, this has been rarely used due to the difficult task of initializing the query vector with a new length (see \ref{sec:model_query_attention}).

Transformer models solve the necessary reduction mainly in two different schemes. In the first scheme, reminiscent of recurrent models, a single token of the last Transformer block is designated to capture and embed the entire sequence in a single token. BERT approached this problem by introducing a \texttt{CLS}-token in the tokenization process and appending it to the input sequence. Similarly, Gao et al. \cite{gao-etal-2021-simcse}, Hou et al. \cite{hou2023improving}, and Wang et al. \cite{wang2023sncse} use a \texttt{CLS}-token to create single token sentence representations and further improve them with contrastive learning schemes. Feng and Yang et al \cite{feng-etal-2022-language} create language agnostic BERT sentence embeddings by using the $\ell_2$ normalized \texttt{CLS}-token of the last encoder block.

The second scheme employs standard pooling algorithms, usually averaging, to combine all hidden word tokens into a single token. 
Sentence-BERT \cite{reimers-gurevych-2019-sentence} uses BERT as a base model in a siamese model setup and creates sentence embedding by averaging over the tokens of the last Transformer block. Refined SBERT \cite{CHU2023126453} extends this approach to a manifold space. Correspondingly, Li et al. \cite{li-etal-2020-sentence} find improved performance averaging over the hidden tokens of the last two Transformer blocks instead of using the \texttt{CLS}-token. Kim et al. \cite{kim-etal-2021-self} follow Li et al. and combine it with a contrastive learning scheme to create sentence embeddings. Barkan et al. \cite{barkan2020scalable} average over the last 4 Transformer blocks and Park et al. \cite{park-lee-2021-finetuning} pool over all hidden encoder embeddings to generate a pooled sentence embedding for the purpose of building a variational autoencoder. Sentence T5 \cite{ni-etal-2022-sentence} employs both approaches in different settings. In an encoder only setting the first token of the last Transformer block is used as an sentence embedding and in an encoder-decoder setting, the sequence is reduced by averaging over all encoder output tokens.
Overall, these two approaches are the current state-of-the-art in reducing sequences down in Transformer models.

Another relevant research area focuses on token pruning. Token pruning aims to avoid unnecessary computation by removing either parameters or word tokens from Transformer. In particular, strategies focused on eliminating word tokens in successive attention layers are of considerable interest. PowerBERT \cite{goyal2020power} computes the importance of each token and only passes on a number of tokens. The importance is taken as the sum of the attention paid to a token by all other tokens in the sequence. In a similar way, length adaptive Transformers \cite{kim-cho-2021-length} prune a random number of tokens, according to importance, after each attention layer. Kim et al. \cite{kim2022learned} avoid the fixed number of tokens pruned by introducing an importance threshold, pruning all tokens falling below this threshold. TR-BERT \cite{ye2021tr} devised a reinforcement learning scheme in which a module is trained to decide which tokens to prune. Meanwhile, Zero-TPrune \cite{wang2023zero} prunes tokens according to importance and similarity to other tokens.
While the goal of reducing the sequence length matches with token pruning models, the motivation for token pruning solely lies in saving computational resources. The attention process itself is not changed and the question of using the sequence reduction as a training tool in attention models remains open. 


Our proposed approach attempts to make use of the attention mechanisms' ability to change the sequence length directly in the attention step. This is done by scaling the query vector generation process with additional parameters. Similar to our approach, Fang et al. create a conditional variational autoencoder by using attention average blocks \cite{fang2021transformer}. These attention average blocks use a single learnable q-vector token in a Multi-head attention step to reduce the sentence down to a single token representations. We extend this approach by allowing the attention process to reduce an arbitrary number of tokens. While introducing strong restrictions, in specific tasks, this new approach will offer a more nuanced way to manipulate the shape of sequences with attention mechanisms.


\section{Model}
\begin{figure*}
\centering
\includegraphics[width=0.8\linewidth]{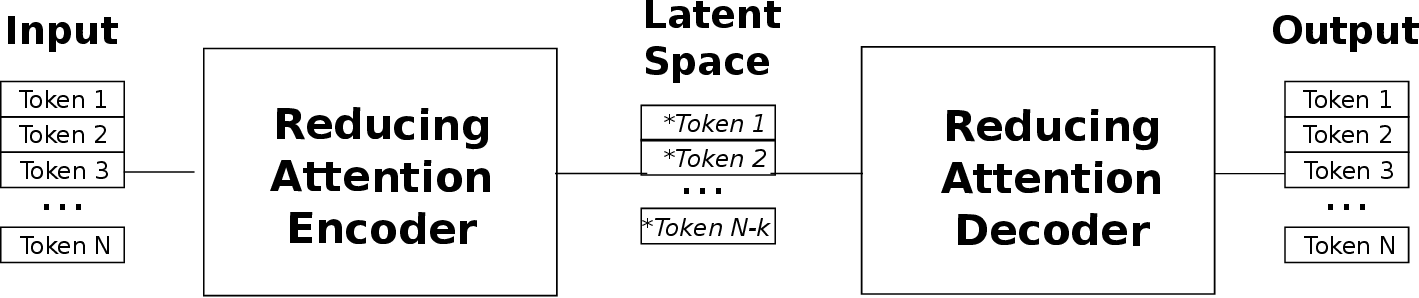}
\caption{\label{fig:model}
Model architecture. The encoder reduces the input sequence from N tokens down to  N-k tokens. This reduced latent space is then given to a decoder to reproduce the original input sequence.
}
\end{figure*}

Our model follows a basic encoder-decoder structure (see Fig. \ref{fig:model}). An input length sequence $N$ is subjected to the attention-based encoder that reduces the sequence by $k$ tokens. The sequence length in latent space is thus reduced to $N-k$ tokens. This reduced latent space is then given to a decoder to reproduce the original input sequence. Thus, the model learns to compress the information of a full-length sequence into a reduced form and then reconstruct the complete sequence from its reduced form.

The reducing attention encoder/decoder blocks are, in essence, applications of the scaled dot-product attention introduced by Vaswani et al. \cite{vaswani2017attention} combined with additional scaling weights to allow for sequence shape manipulation. We chose this form of attention as it is well established in existing Transformer models, and it fundamentally already allows for sequence manipulation.

\subsection{Attention can manipulate sequence length directly}
\label{sec:model_query_attention}

Looking at how the scaled dot-product is calculated immediately makes apparent how it allows for direct manipulation of the sequence shape. By generating query, key and value vectors $q ,k ,v$ from original input $x$, the attention $\mathcal{A}$ is calculated via the formula:
\begin{equation}
\label{eq:attn}
\mathcal{A}(Q, K, V) =  \mathrm{softmax} ( \frac{QK^T}{\sqrt{d_k}} ) V \\
\end{equation}
where $Q, K, V$ are the respective matrices consisting of the query, key, and value vectors for each token in the sequence. Looking at the dimensions of the variables in the formula clarifies how the sequence shape is directly manipulated. Considering an input sequence with $n$ tokens and word embedding dimension of $d_m$, the input sequence matrix $X$ has the dimension of $\mathds{R}^{n \times d_m}$. Multiplying $X$ with the respective weight matrix $W^{Q,K,V}$ generates the respective query, key and value matrices:
\begin{equation}
\label{eq:qkv}
Q, K, V = X \times W^{Q,K,V}
\end{equation}
The dimension of the $Q, K, V$ matrices are then given by
\begin{equation*}
\mathds{R}^{n \times d_m} \times \mathds{R}^{d_m \times d_{q,k,v}} = \mathds{R}^{n \times d_{q,k,v}}
\end{equation*}
with the respective vector dimension $d_q, d_k, d_v$ that can be arbitrarily chosen, whereas the dimension of the query and key vectors need to match $d_q=d_k$. It is a common convention in Transformer models to set the query, key, and value to the same value $d_q=d_k=d_v$ as was done in the introducing Transformer paper. We can see that the sequence length $n$ is carried over from the original input and is equal for all matrices. When artificially labeling the sequence length of the query, key and value matrices, we get the respective dimension $\mathds{R}^{n_{q,k,v} \times d_{q,k,v}}$. Plugging the Q, K and V matrices into equation \ref{eq:attn} and looking at the dimensions, it becomes apparent that the dimension of the scaled dot-product has the form
\begin{equation*}
\mathds{R}^{n_q \times d_v}
\end{equation*}
This shows that the output sequence shape is only dictated by the number of tokens in the query matrix. Thus, the sequence shape can be directly manipulated if the number of tokens in the query matrix differs from the original sequence length, while the key and value matrices still carry the full information of the complete input sequence.
Consequentially, the difficulty of manipulating the sequence shape lies in initializing the query vector matrix with a different shape than its key and value counterparts ($n \neq n_q$).

\subsection{Initializing query vector with addition scaling matrix}
The simplest method to initialize the query vectors is by introducing an additional scaling matrix of trainable parameters $W^{S}$, with the dimensions $\mathds{R}^{n_q \times n}$, where $n_q$ and $n$ denote the new and old sequence length, respectively. Inserting $W^{S}$ into equation \ref{eq:qkv} yields a query matrix of the desired shape.
\begin{equation*}
Q = W^S \times (X \times W^Q)
\end{equation*}
with
\begin{equation*}
\mathds{R}^{n_q \times n} \times \mathds{R}^{n \times d_m} \times \mathds{R}^{d_m \times d_q} = \mathds{R}^{n_q \times d_q}
\end{equation*}
Thus, the original sequence length $n$ given by the input $X$ has been changed to the new sequence length $n_q$, where $n_q$ is an arbitrarily chosen value.

\subsection{Method limitations}
\label{sec:method_limits}
This scaling matrix method immediately exposes some of the constraints and challenges of this sequence shape-changing approach.

While neural networks need the input to have the same dimension on the token level, they are more flexible regarding sequence length. Adding the scaling matrix $W^S$ negates this flexibility and fixates the input and output sequence lengths allowed to a specific value. This additional strong constraint is undesirable in many ways. Sequences, especially in NLP, seldom have the same length across the entire dataset. Thus, all input sequences presented to the model will need to be padded or constrained to the same length, which introduces additional preprocessing and computational costs due to the increased padding. Additionally, if we have to pad sequences with more tokens that we effectively reduce, in an ideal case, this model would then remove the padding in the sequence that we had to add to use the model.

For instance, consider a scenario where we pad all sequences in a dataset to match the length of the longest one and then train the model to decrease this sequence length by 50\%. If the dataset's average sequence length is less than half of the longest sequence's length, we will end up incorporating more padding tokens than the number of tokens we're reducing, on average. This will likely impact the model's desired ability to identify and compress information-carrying features in the sequence, as it will mainly reduce padding tokens void of information.

This strong additional constraint makes this model undesirable for tasks that handle short sequences and sequences of varying lengths. In cases where the length of sequences is already fixed or the sequences' lengths are similar, the input length constraint this model introduces has less of a negative impact. We argue that this constraint will effectively have no negative consequence on tasks considering long sequences. Many state-of-the-art Transformer models already have an input limit of 512 to 728 tokens due to the associated computational cost of longer sequences. This already establishes a constraint for the input length, and our model's fixed input length has no additional effect.

Consequentially, NLP tasks where the input sequence length exceeds the transformer-imposed input length limit of 512 tokens can benefit from this model, even with the strong constraints. Part of future research will be to investigate methods to make the initialization of the query matrix more flexible to enable the model to handle shorter sequences better. Nevertheless, many tasks and use-cases could benefit from this model approach, such as long document classification.

\section{Experiments}

In our experiments, we employ the proposed method in an autoencoder setting. Our goal is to investigate the models' ability and limits regarding their ability to retain most of the important information of a sequence while reducing the length of the sequence. Thus, we define the performance of the model as the models' ability to reconstruct the the original sequence from its latent space representation. If the model can reconstruct the original sequence from its reduced form, we argue that the reduced form has retained all the information of the original full-size sequence. If the model cannot reproduce the original sequence fully, it can be surmised that some critical information has been lost in the reduction process. 

The goal of our investigation is to study and explore the limits of the reduction process. How far can we reduce a sequence without it losing critical information? How does the models performance depend on the input sequence length and the desired length of latent space?

\subsection{Hyperparameters and Setup}
To answer these questions, we train our model in several different input and latent sequence length settings. In the first experiment, we fixate the input sequence length and systematically reduce the latent sequence length. This allows us to investigate the models learning behaviour for different reduction sizes and potential limits in how much a sequence can be reduced in this setting.
For the second experiment we repeat the first experiment for different input sequence lengths. This explores how the model is affected by different input sequence lengths and whether the input sequence length has an influence on how much we can reduce it without losing information.

Table \ref{tab:hyperparams} list the hyperparameters and settings used in this study. The model was trained for a maximum of 20 epochs with a patience of 5 epochs. The performance of the model is recorded as the token-wise accuracy between the original sequence and the reconstructed sequence. Thus, an accuracy of $1.0$ indicates that every token in the entire sequence was correctly reconstructed, while an accuracy of $0.5$ would indicate that only half of the tokens in the sequence were correctly decoded.
The model was trained using cross entopy loss and the learning rate was linearly reduced from its start value of $0.001$ to $0.0001$ over the first five epochs. For input sequences above 256 tokens, the learning rate was kept static at $0.0001$.
The chosen hyperparameter settings are the result of a hyperparameter search for the case of reducing a 512 input token sequence down to a 256 latent token sequence. Initially, a dropout of 30\% was considered during the hyperparameter search it became apparrent that not adding any dropout yielded the best results for longer sequences.

\begin{table} [h]
    \label{tab:hyperparams}
    \centering
    \caption{Hyperparameter Setup}
    \begin{tabularx}{\linewidth}{rX}
    \toprule
    \textbf{Parameter} & \textbf{Value/Description} \\
    \midrule
    Optimizer & AdamW \\
    Loss Function & CrossEntropyLoss \\
    Learning Rate & Start: $0.001$/ End: $0.0001$\\
    Number of Epochs & 20 \\
    Patience & 5 \\
    Embedding dimension & 256 \\
    Attention dimension & 512 \\
    Batch size & 16 \\
    \bottomrule
    \end{tabularx}
\end{table}

\subsection{Dataset}
The model was trained with data from the wikipedia dataset \cite{wikidump}. Throughout this study we used data from the English '20220301.en' data dump. As the model has to be trained many times thoughout our investigations, as a time saving feature, we only used the first 30\% of the dataset in our simulations. This leaves us with a dataset consisting of roughly $1.9$ million samples which were further divided into an 80/20 train/test split. Each sample in the dataset corresponds to the contents of a single wikipedia page. To generate samples of the desired input sequence length, we only used the first $N$ tokens of the repsective wikipedia page. Thus, we create a single sample per page and we observe the same number of data samples for different input sequence lengths.
The data was tokenized via the Transformers library, using the pretrained BertTokenizer \cite{wolf-etal-2020-transformers}.

\section{Results and Discussion}

\subsection{Impact of Sequence Reduction on Model Accuracy}

\begin{figure}
\centering
\includegraphics*[width=\linewidth]{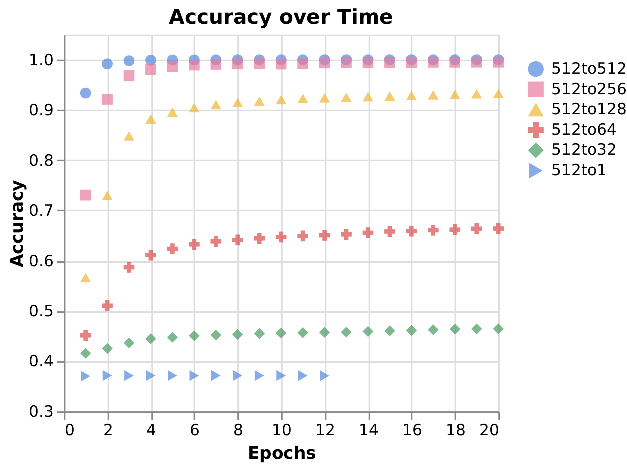}
\caption{\label{fig:pot}
Reconstruction accuracy over training epochs for different latent sequence lengths. The input sequence length was fixed to 512 tokens.}
\end{figure}

To start testing our hypothesis, we fix the input sequence length to 512 tokens and we train the model to reduce it down to different latent sequence lengths. The results of these simulations are shown in Figure \ref{fig:pot}. The graph depicts the recorded validation accuracy over training epochs.

\begin{figure}
\centering
\includegraphics*[width=\linewidth]{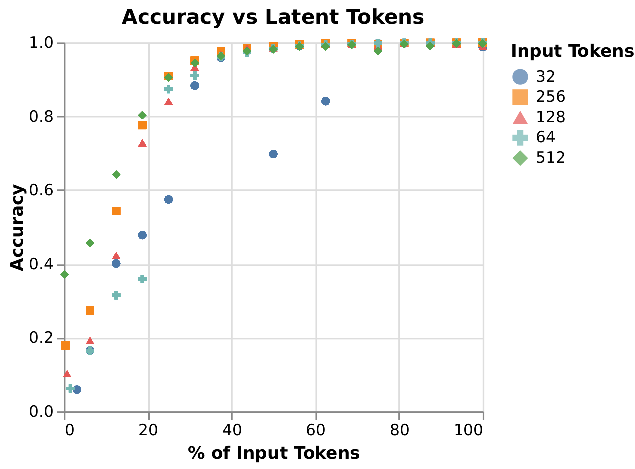}
\caption{\label{fig:pol}
Reconstruction accuracy over latent sequence lengths expressed as a percentage of respective number of tokens in the input sequence length.}
\end{figure}

As expected, when not reducing the input sequence at all (\textit{512to512)} the model is able to reproduce the original sequence completely. More interesting is the fact that the model continues to show this near-perfect performance when the model reduces the original sequence by half (\textit{512to256}). Even when reducing the original sequence down to $25\%$ of its original length (\textit{512to128)}, the model still reaches a reconstruction accuracy of over $90\%$. The performance of the model starts to fall more significantly when approaching reduction ratios of $20\%$ or lower (see Fig. \ref{fig:pol}).

These results prove that in the case of an input sequence length of 512 tokens, we can reduce the sequence length by half without losing any critical information. Moreover, most of the critical information are retained even when the original sequence is reduced down to $25\%$. \\

\begin{figure}
\centering
\includegraphics*[width=\linewidth]{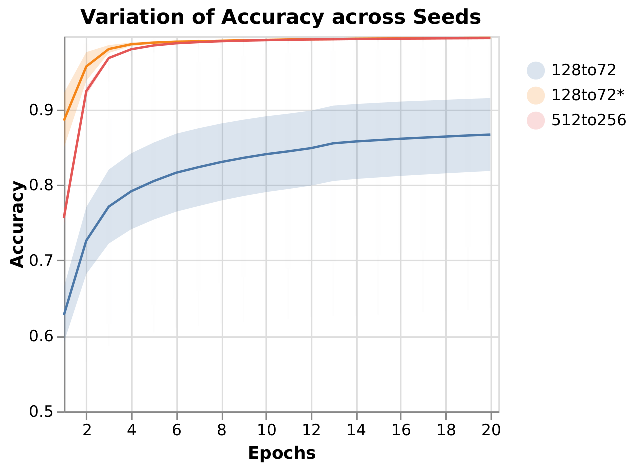}
\caption{\label{fig:errorband}
Illustration of accuracy variance for different input length sizes. Depicted are errorbands for three different simulation runs. The bold line depicts the mean value. The simulation denoted with $*$ ran with an increased learning rate of $0.001$ and linearly decreased, in 5 epochs, to $0.0001$. $0.0001$ is the default learning rate.}
\end{figure}

\subsection{Impact of Input Sequence Length on Model Accuracy}

To further understand the behaviour of our model we then focus on the impact of different Input Sequence Lengths on model accuracy. Figure \ref{fig:pol} illustrates how the model performs with varying degrees of input sequence reduction. To allow a more direct and easy comparison between simulations with different input sequence lengths, the x-axis expresses the latent sequence length as a percentage of the corresponding input.

In Figure \ref{fig:pol}, focusing on the case of 512 input tokens, we can observe the previously described behaviour. 
Up to a $50\%$ reduction, the model achieves near-perfect accuracy. However, greater reductions result in a noticeable decline in performance. The model behaves differently with shorter input sequences: performance degrades more quickly for shorter inputs compared to longer ones. Simulations with inputs longer than 256 tokens maintain performance above a $0.9$ accuracy level even at a $25\%$ reduction, whereas those with shorter inputs fall below this threshold.
In particular, when looking at the simulations below a $64$ input sequence length, we observe stronger fluctuations in performance.

The cause of this increased drop in performance for smaller input sequence lengths is not immediately evident. 
One possible explanation could be the nature of the sequences as natural text, which has inherent structural rules. Longer sequences may have more of these rules and repetitive elements, potentially making them easier to reduce efficiently. For instance, a paragraph made up of multiple sentences will exhibit repeated grammatical structures and commonly used words. Such redundancy in longer sequences might be more effectively compressed by the model in the latent space compared to shorter sequences.

Another potential reason could be the quantized nature of our sequences. A single token carries more weight in a 64-token sequence than in a 512-token sequence. Despite examining proportional reductions in our study, the effect of removing a single token could be more impactful in models with shorter input lengths.

Finally, the decline in performance might also be attributed to the fewer latent tokens available in models with shorter input sequences. For instance, a $25\%$  reduction from 64 tokens leaves only 16 latent tokens, while the same reduction from 512 tokens yields 128 latent tokens. Although larger sequences require encoding more information, the greater number of latent tokens could allow the model to learn and capture more nuanced relationships.

Overall, we can summarize that for all observed input sequence lengths, we can at least halve the sequence in latent space without loss of significant information. In a simplified view, we could say that each latent space token is able to carry the information of two input tokens without losing information. However, attempting to condense the information of three tokens into one introduces the first small but significant loss in information. When reducing a sequence of $30\%$ or less, we retain an accuracy of above $0.9$. However, reductions bigger than $30\%$ lead to a substantial loss of information, the extent of which varies depending on the initial number of input tokens.

\begin{figure*}
\centering
\includegraphics*[width=\linewidth]{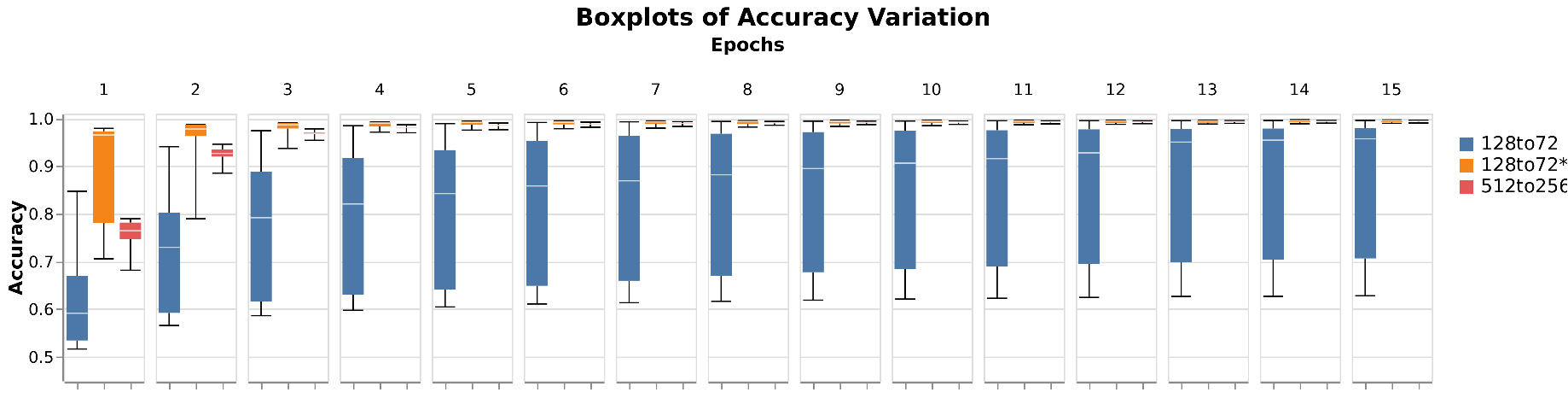}
\caption{\label{fig:boxplot}
Illustration of accuracy variance for different input length sizes via box plots. Box plots for three simulations are compared over 20 epochs. The simulation ran with a learning rate of $0.0001$, the simulation denoted with $*$ ran with an increased learning rate of $0.001$ and linearly decreased, in 5 epochs, to $0.0001$.}
\end{figure*}

\subsection{Performance variance due to initialization}

In initial simulation runs, we used a fixed learning rate of $0.0001$ for all simulations. While performing well for large input sequence lengths, when scaling down to smaller input sequence lengths we could observe a large variance in performance depending on the initialization of the model itself. To further examine this behaviour we simulated three different input/latent sequence length combinations $10$ times with different initialization seeds. Figure \ref{fig:errorband} shows the results of these simulations reporting the accuracy variance over the training epochs. The error-bands in the graph represent the standard error.

Looking at the curve for the larger input sequence length, we see a very small variance in accuracy after training. Thus, in the case of the large input sequence, the model was able to consistently reach a high performance. In contrast, the simulation with an input sequence length of $128$ shows a significantly larger variance in accuracy. A more detailed quantification of model behavior is presented in Figure \ref{fig:boxplot}, which shows boxplots for accuracy variation across each epoch. These boxplots capture the range of achieved accuracy over $10$ runs, with the upper and lower bounds representing the maximum and minimum accuracies attained. Looking at the accuracy after 20 epochs of training, the accuracy averages around $0.9$ for $128$-token input sequence length. The arms of the boxplot further indicate that the minimum and maximum reached accuracies range from close to $1$ to $0.7$. This large discrepancy in performance between different initializations suggests that in some cases the model got stuck in local minima and did not reach the possible global minimum. This problem was alleviated by increasing the initial learning rate by a factor of $10$ and linearly decreasing it over the next 5 epochs down to its original value of $0.0001$. The simulation called \textit{128to72*} depicts the results of this learning rate approach. We can see that the model still shows a higher variability after the first epoch but quickly finds the correct global minimum in subsequent epochs.  Overall, the learning rate was the only parameter that had an effect on this problem. Making the model more complex by increasing the embedding or attention dimension did not affect the variance for smaller input sequence lengths.

We hypothesize that altering the input sequence length affects the smoothness of the error function, leading to more distinct local minima that are challenging to circumvent.

\section{Conclusion}
In this study, we introduced a new way to use the well-established scaled-dot product attention of the Transformer to directly manipulate the shape of sequences. The scaled dot-product directly allows the change of the length of a sequence by adding a scaling matrix to the query vector generation.
We have developed a new autoencoder that compresses the original input sequence into a shorter latent sequence and then reconstructs the original input from this compressed latent space. In current literature, no previous work has looked at using sequence length manipulation as an encoding tool in attention-based deep learning networks.

Our subsequent exploration of the models' capabilities reveals interesting insights into the model's reduction behaviour. We were able to observe that in latent space, the original sequence length can be reduced by half without information being lost. Reducing the original sequence down to $30\%$ of its original length still allowed for a reconstruction accuracy of over $90\%$.
We could further observe that the accuracy starts to drop faster for smaller input sequence lengths. We hypothesised that this is most likely due to a combination of the facts that large natural text carries a lot of recurring structural and word information and each token carries more weight and has more impact in smaller sequences.

We further were able to observe that the model is more prone to get stuck in local minima for smaller input sequence lengths. Increasing the learning rate proved to alleviate this effect. Thus, choosing the appropriate learning rate for the individual use case is of great importance. We could observe that larger input sequence length models were able to perform well with lower learning rates than their smaller counterparts.

One of the larger drawbacks of this model is the introduction of an additional strong restriction regarding the possible shape of the input of the model. In its current form, the model requires the input shape to be static. This means, the model is not able to process sequences of different lengths as most NLP models are able to do. This might prove problematic in NLP setting where sequences seldomly are of the same length. 

Nevertheless, we argue that in the field of NLP there are use cases and tasks where this additional restriction has little to no effect. This is further supported by the already existing and common upper limit of $512$ and $768$ token sequences in Transformer models.

This study opens up exciting further investigations. Applying this model to sequences outside of the NLP setting would allow to study whether the faster dropoff in performance for smaller input sequence lengths is due to the specific structure of natural text sequences or due to other effects.
A further study into making the model more flexible regarding the input sequence length would make this model more interesting for tasks with highly varying sequence lengths.
Furthermore, we investigated the ability to reduce sequences in a single attention step. A promising next step would be to investigate the reduction limits when considering multiple attention steps.

Ultimately, we believe that the direct and more nuance manipulation of sequence lengths is an overlooked avenue in tuning and designing machine learning models, especially in cases where sequential data needs to be reduced down to one vector.


\bibliographystyle{IEEEtran}
\bibliography{references}


%
%




\end{document}